\documentclass{esannV2}
\usepackage{graphicx}
\usepackage[latin1]{inputenc}
\usepackage{amssymb,amsmath,array}
\usepackage{subcaption}
\usepackage{color}
\begin{document}
\title{Vector Field Based Neural Networks}

\author{Daniel Vieira$^1$, Fabio Rangel$^1$ , Fabr\'icio Firmino$^1$ and Joao Paixao$^1$
%
\thanks{The authors would like to acknowledge CNPq/CAPES for funding.}
%
\vspace{.3cm}\\
%
1- Universidade Federal do Rio de Janeiro (UFRJ)\\
Graduation Program in Informatics (PPGI) \\
Av. Athos da Silveira Ramos, 149, Rio de Janeiro, RJ - Brazil
%
}

\maketitle

\begin{abstract}
A novel Neural Network architecture is proposed using the mathematically and physically rich idea of vector fields as hidden layers to perform nonlinear transformations in the data. The data points are interpreted as particles moving along a flow defined by the vector field which intuitively represents the desired movement to enable classification. The architecture moves the data points from their original configuration to a new one following the streamlines of the vector field with the objective of achieving a final configuration where classes are separable. An optimization problem is solved through gradient descent to learn this vector field.

\end{abstract}

\section{Introduction}

Understanding how \textit{Black Box} models \cite{freitas2010importance} work, such as Deep Neural Architectures \cite{goyal2016interpreting} and Support Vector Machines \cite{barakat2006rule}, is an extensively discussed problem in the literature \cite{Freitas2014}, where many works focus in visualizing and understanding their behavior \cite{zeiler2014visualizing} \cite{yosinskiunderstanding} \cite{olah2017feature}. It is possible to address the problem of comprehending Neural Networks behavior in a geometrical sense by understanding them as universal function approximators \cite{hornik1991approximation}. Another possible interpretation considers the hidden layers in Neural Networks as nonlinear continuous transformations of the domain space \cite{colah}. Towards the end of his work, the author suggests that vector fields might be better to handle these transformations than traditional layers. Inspired by \cite{colah}, the present work proposes to combine Neural Networks with vector fields in order to understand data separation with data points as particles moving along a flow.

Vector fields have also been recently used to analyze the optimization problem in Generative Adversarial Networks (GANs) \cite{mescheder2017numerics}, achieving remarkable results for visualization and comprehension of GANs limitations and how to extend them. This work introduces a novel architecture using vector fields as activation functions. We optimize vector fields parameters through stochastic gradient descent using binary cross entropy as loss (cost) function.


By applying the concept of vector fields in Neural Networks, a vast amount of established mathematical and physical concepts, abstractions, and visualizations arises for the Neural Networks. For instance, Euler's method for solving ordinary differential equations \cite{Butcher1987} is used in this work to implement the concept of data points as particles moving through a flow. 

This paper presents a computational experiment performed over three nonlinear separable, two dimensional datasets, using a vector field generated by a simple Gaussian kernel function. With different initialization hyperparameters the cost function shows consistent reduction through epochs and the results are further analyzed. Section \ref{vector_field_theoretical} describes the architecture and the optimization problem.


\section{Vector Field Neural Networks} \label{vector_field_theoretical}

A vector field on $\mathbb{R}^{n}$ is a smooth function $K: \mathbb{R}^{n} \to \mathbb{R}^{n}$. Consider the corresponding ordinary differential equation (ODE):
$$ X'(t) = K(X(t)), \  X(t_0)=X_0$$

with $X \in \mathbb{R}^{n}$. A curve $X(t)$ solving the ODE is called an streamline of the vector field $K$. Given a particle in a position $X(t_0)=X_0$ at time $t_0$, a possible physical interpretation is where each vector $K(X)$ represents the velocity acting on the particle at a given point in space and the streamline is the displacement done on the particle as it travels along the path $X(t)$. At time $t_N>t_0$, the particle will be in position $X(t_N)$.

Given a family of vector fields $K(X,\theta)$ defined by some parameters $\theta$, we proposed to find the best vector field in this family aiming to transform every point $X_0$ in the input space, in a point  $X(t_N)$ in the transformed space such that points in distinct classes would be linearly separable. Intuitively, the vector field represents the desired movement to enable the classification. 




Euler's method \cite{Butcher1987} is used to approximate $X(t_N)$ by $X_N$, the solution of the ODE, with $X_i \approx X(t_0 +ih)$ for the discretization and $K(X, \theta)$ as our vector field with the iteration:

\begin{equation} \label{eq:101}
X_{i+1} = X_i + hK(X_i,\theta)  \qquad 0\leqslant i\leqslant N,
\end{equation}

where $h$ (the step size) and $N$ (number of steps), such that $t_N=t_0+Nh$, are hyperparameters and $\theta$ represents the vector field parameters. Considering Euler's method,  it is known that for $h \to 0$, the streamlines of $K(\theta, X)$ are calculated exactly. 

\begin{figure}[!ht]
     \centering
     \begin{subfigure}[b]{0.28\linewidth}
         \includegraphics[width=\linewidth]{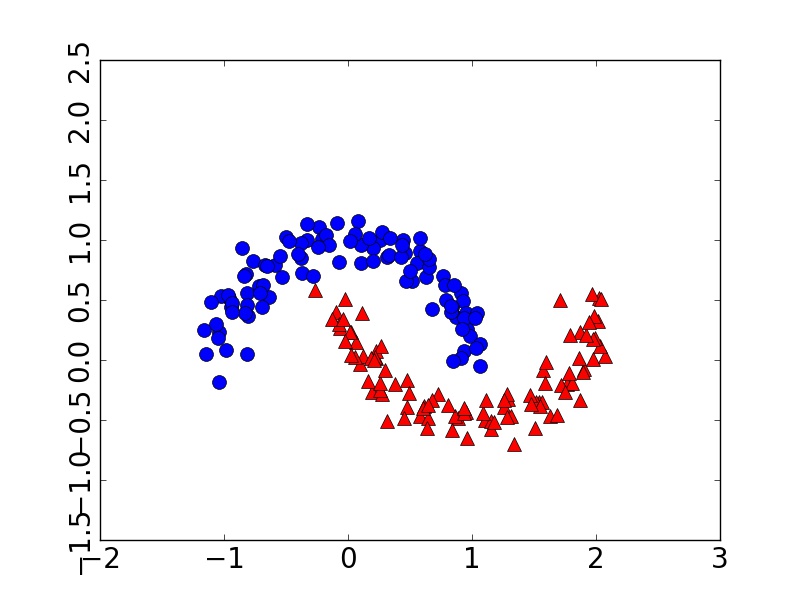}
         \label{fig:original}
     \end{subfigure}
     ~ 
     \begin{subfigure}[b]{0.4\linewidth}
         \includegraphics[width=\linewidth]{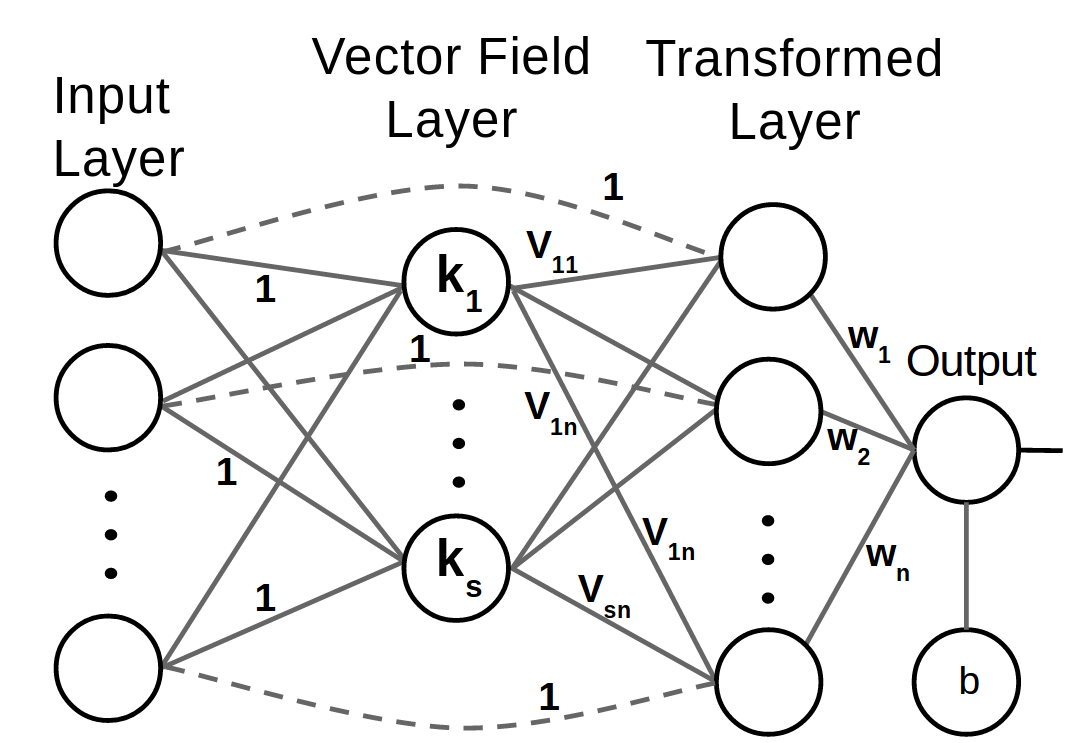}
         \label{fig:architecture}
     \end{subfigure}
         \begin{subfigure}[b]{0.28\linewidth}
         \includegraphics[width=\linewidth]{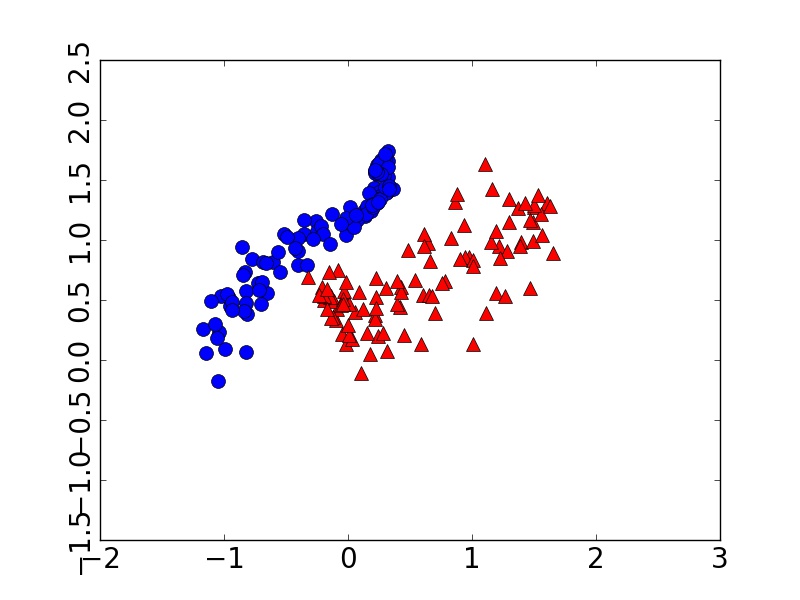}
         \label{fig:transformed}
     \end{subfigure}
     \begin{subfigure}[b]{0.3\linewidth}
         \includegraphics[width=\linewidth]{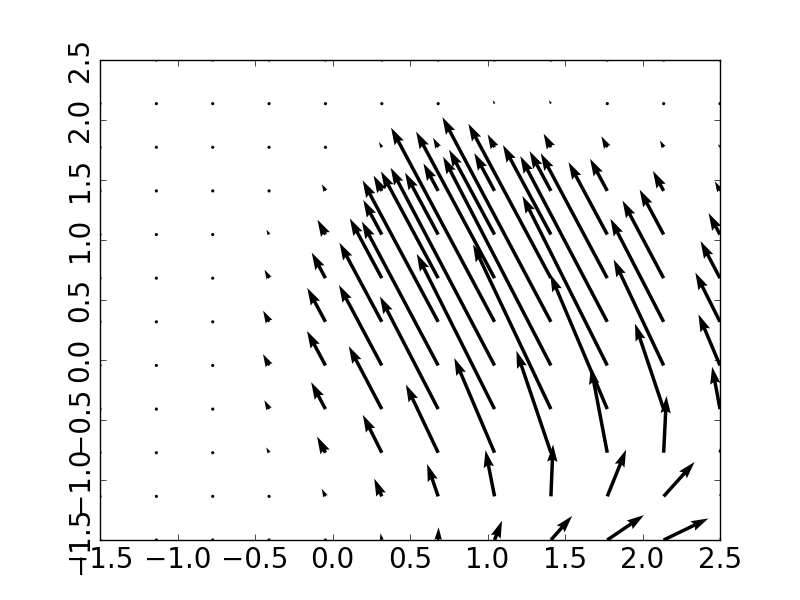}
         \label{fig:field}
     \end{subfigure}
         \begin{subfigure}[b]{0.3\linewidth}
         \includegraphics[width=\linewidth]{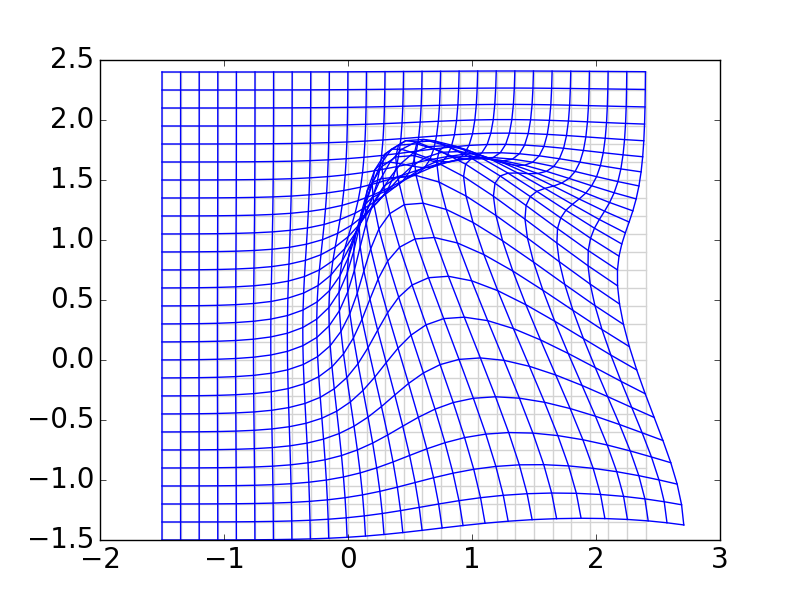}
         \label{fig:meshgrid}
     \end{subfigure}
     \vspace{-0.25cm}
     \caption{From left to right, first row presents input data, the architecture, and the transformed data by the vector field layer. Second row presents the vector field and the space distortion.}
     \label{intro}
 \end{figure}

Figure \ref{intro} presents the input data being transformed by the vector field layer of the architecture. It also presents the optimized vector field which its goal is to linearly separate the data. Note that the architecture's last layer is a linear separator, which can be implemented using a logistic function.

\section{Methodology}

\subsection{Vector Field Optimization}
A computational experiment was developed using a simple kernel function to define the family of vector fields. Kernel $K(X,\theta)$ is presented in Equations \ref{kernel1} and \ref{kernel2}, where $V = \{V_1, V_2, \ldots, V_S\}$ are the vectors where $V_i \in  \mathbb{R}^{n}$ and $M=\{\mu_1, \mu_2, \ldots \mu_S \}$ are the means of the Gaussians  where $\mu_i \in  \mathbb{R}^{n}$, $\theta=\{V,M\}$, and $S$ is the number of Gaussians,
\begin{equation} \label{kernel1}
G(X,\mu) = e^{-||X-\mu||^2},       \qquad X,\ \mu \in \mathbb{R}^{n},
\end{equation}
\begin{equation}\label{kernel2}
 K(X,\theta) = \sum_{i =1}^{S}V_i G(X,\mu_i), \qquad  V_i \in \mathbb{R}^{n}.
\end{equation}
Note that variance is controlled through the $V_i$ parameters, which can be thought of as weight vectors that provide a direction to the vector field (see Figure \ref{intro} for an example). The final layer is a logistic function with a binary cross entropy cost that acts upon the transformed points $X_N$:
\begin{equation}
\hat{y} = \frac{1}{1 + e^{-(w^t X_N +b)}} ,\qquad w,\ X_N \in \mathbb{R}^{n} \ and \ b, \ \hat{y}\in\mathbb{R} \ st. \ 0<\hat{y}<1.
\end{equation}
Let $w$, $V_i$ and $\mu_i$ initial values be random vectors with distribution $U[0,1]$ in $\mathbb{R}^{n}$ and $G_i$ the corresponding distribution and L2 regularization of vectors $V_i$. The gradient of $\mu_{j,i}$ and $V_{j,i}$,  the respective $j$-th components of vectors $\mu_i$ and $V_i$, are presented in Equations \ref{grad1} and \ref{grad2}, where $x_j$ is the $j$-th component of $X$, $w_j/w_k$ are the $j$-th and $k$-th components of vector $w$ and $\lambda$ is the regularization parameter. 
\begin{equation}\label{grad1}
\frac{\partial{C}}{\partial{\mu_{j,i}}} = 2(\hat{y} - y)G(X , \mu_i)(x_j - \mu_{j,i})
\sum_{\substack{k=1} }^{n}w_k V_{k,i},
\end{equation}
\begin{equation}\label{grad2}
\frac{\partial{C}}{\partial{V_{j,i}}} = (\hat{y} - y)G(X, \mu_i)w_j + \eta\lambda V_{j,i}.
\end{equation}
\subsection{Experiment Design}

\begin{figure}[!ht]
    \centering
    \begin{subfigure}[b]{0.31\textwidth}
        \includegraphics[width=\textwidth]{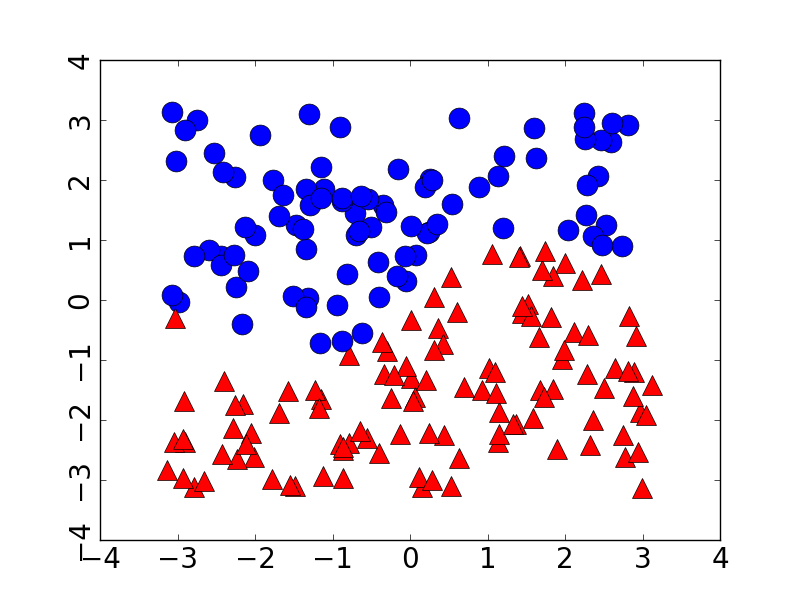}
        \label{fig:sin}
    \end{subfigure}
    ~ 
    \begin{subfigure}[b]{0.31\textwidth}
        \includegraphics[width=\textwidth]{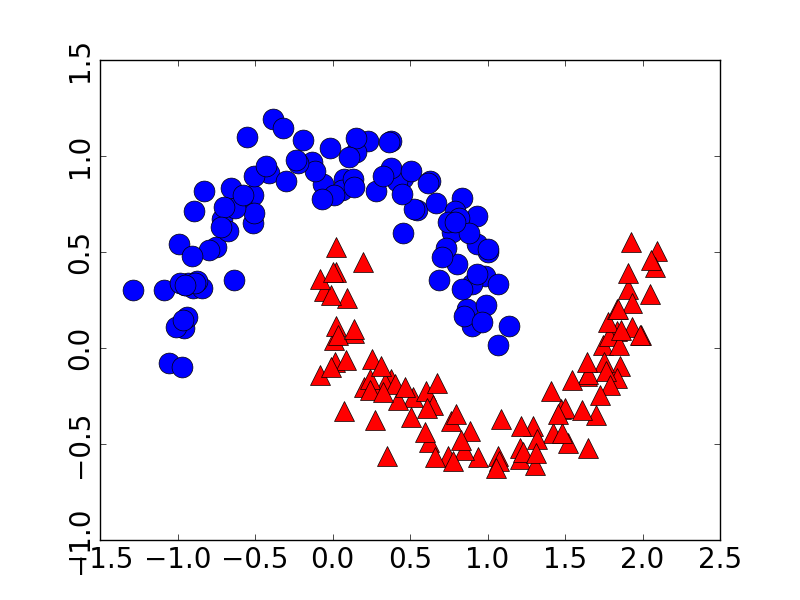}
        \label{fig:moons}
    \end{subfigure}
    ~ 
    \begin{subfigure}[b]{0.31\textwidth}
        \includegraphics[width=\textwidth]{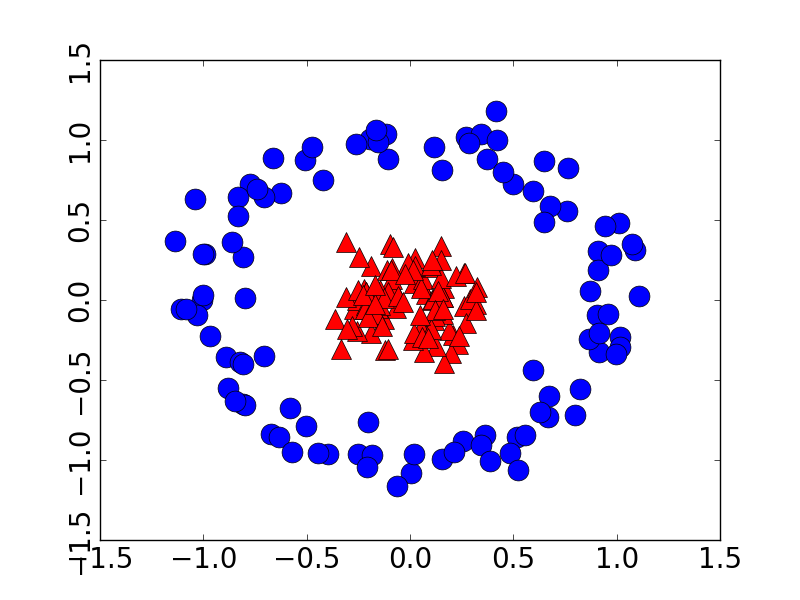}
        \label{fig:circles}
    \end{subfigure}
    \vspace{-0.25cm}
    \caption{\vspace{-0.1cm} Sin, Moons and Circle datasets, respectively.}\label{fig:datasets}
\end{figure}

Two \textit{scikit-learn} machine learning datasets \cite{scikit-learn} (moons and circle) and a sin dataset (created by the authors) were used in this paper. Figure \ref{fig:datasets} shows the datasets.

 
For each dataset, the cost function is evaluated 30 times, using random initial values for the $\theta$ parameters, learning rates in $\{0.03, 0.30, 3.00\}$, and no regularization ($\lambda = 0$). Hyperparameters $h$ and $N$ are both set to one and $S=2$ in this paper's experiments. Training is made with the entire dataset in a full-batch fashion, hence, there is no validation/test set. Results are presented for the cost throughout 10000 epochs in the \textit{circles} dataset. Next, the boundary layer is calculated as a color map, with the original data points plotted over it. Thus, we can analyze the relationship between the boundary layer in the original and transformed space. Finally, the effects of regularization can be visualized and compared in a color map for \textit{sin} dataset.


\section{Results and Discussion}

 \begin{figure}[!htb]
     \centering
     \begin{subfigure}[b]{0.31\textwidth}
         \includegraphics[width=\textwidth]{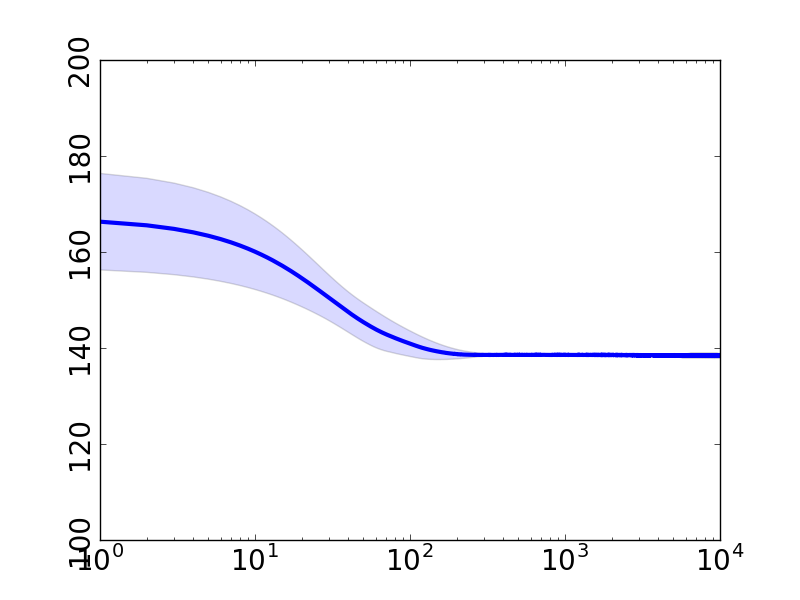}
         \label{fig:cost003}
     \end{subfigure}
     ~ 
     \begin{subfigure}[b]{0.31\textwidth}
         \includegraphics[width=\textwidth]{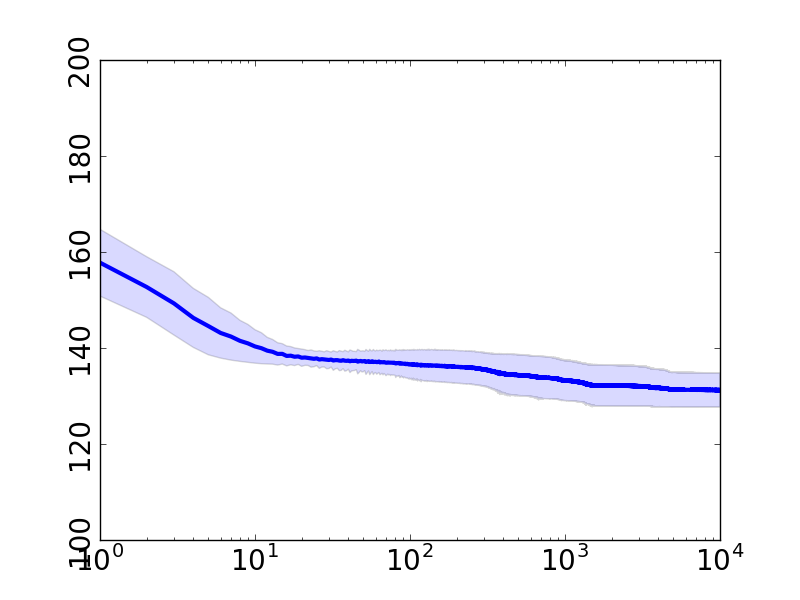}
         \label{fig:cost030}
     \end{subfigure}
     ~ 
     \begin{subfigure}[b]{0.31\textwidth}
         \includegraphics[width=\textwidth]{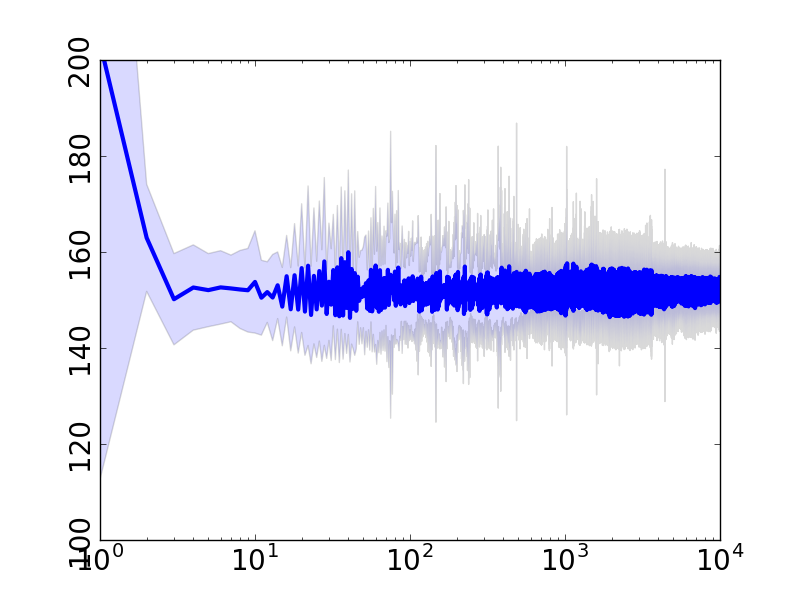}
         \label{fig:cost300}
     \end{subfigure}
    \vspace{-0.25cm}
     \caption{\vspace{-0.1cm}Cost vs. Epochs. \textit{Circles} dataset with $\theta = \{0.03, 0.3, 3.0\}$ respectively.}\label{fig:cost}
 \end{figure}

Analyzing the cost function along the epochs for different learning rates for the \textit{circles} dataset shows the reduction of cost through epochs and an interesting pattern appears: as the learning rate increases the cost function and its deviation becomes less smooth and the standard deviation increases as well.

\begin{figure}[!htb]
    \centering
    \begin{subfigure}[b]{0.3\textwidth}
        \includegraphics[width=\textwidth]{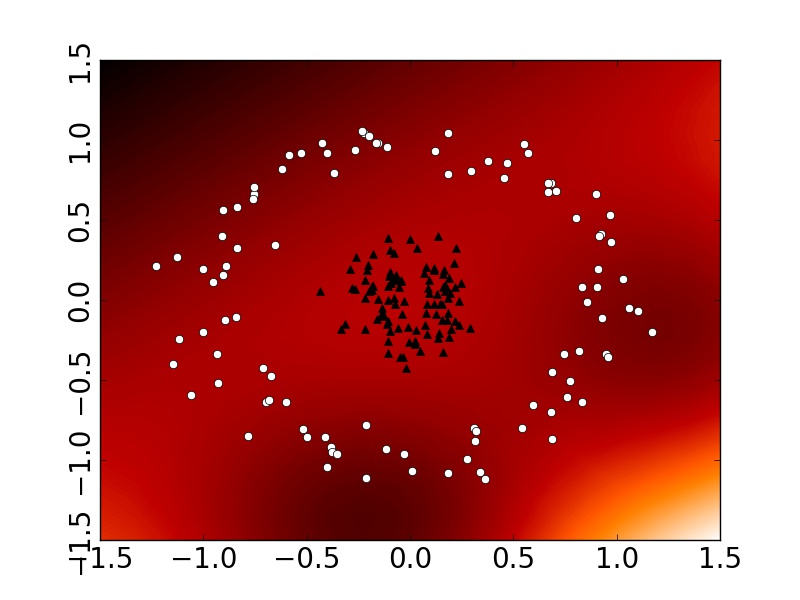}
        \label{fig:boundo030}
    \end{subfigure}
    ~ 
    \begin{subfigure}[b]{0.3\textwidth}
        \includegraphics[width=\textwidth]{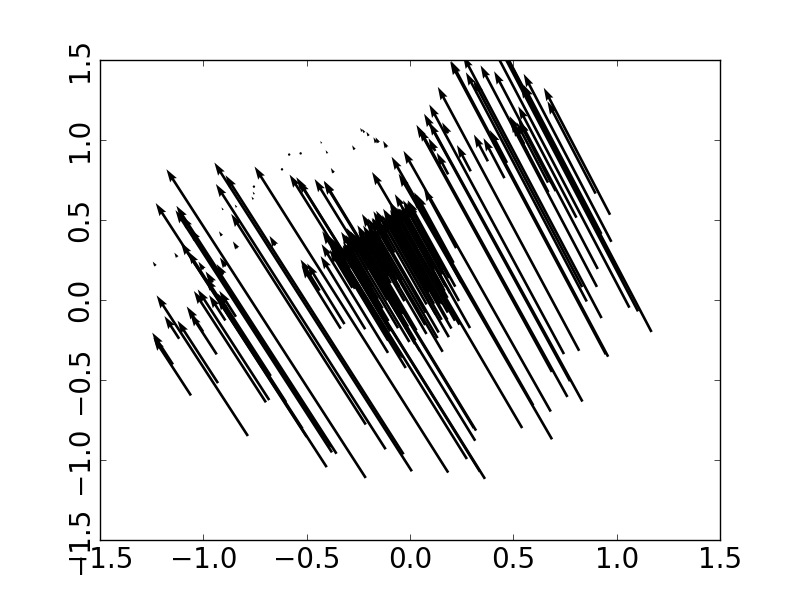}
        \label{fig:vec030}
    \end{subfigure}
    ~ 
    \begin{subfigure}[b]{0.3\textwidth}
        \includegraphics[width=\textwidth]{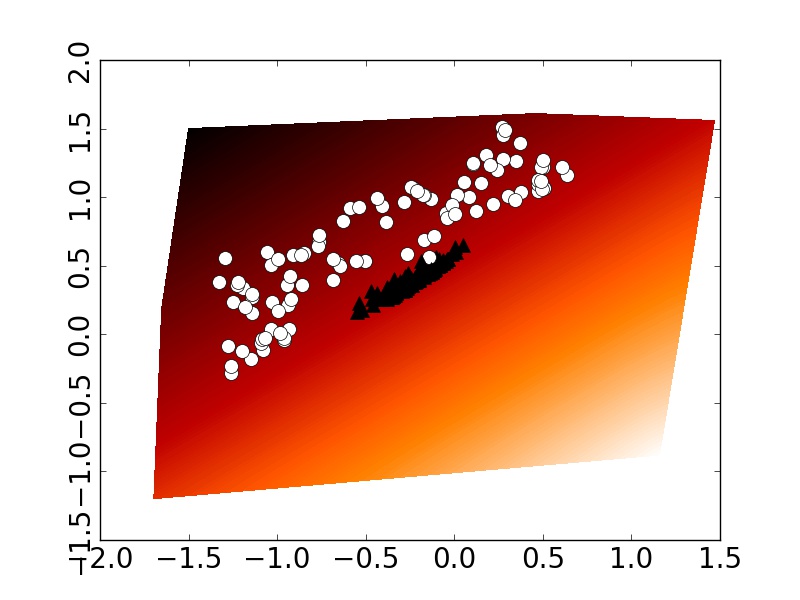}
        \label{fig:boundt030}
    \end{subfigure}
    \vspace{-0.25cm}
    \caption{Original space; vector field; and transformed space, respectively. }\label{fig:bound}
\end{figure}

\begin{figure}[!htb]
    \centering
    \begin{subfigure}[b]{0.34\textwidth}
        \includegraphics[width=\textwidth]{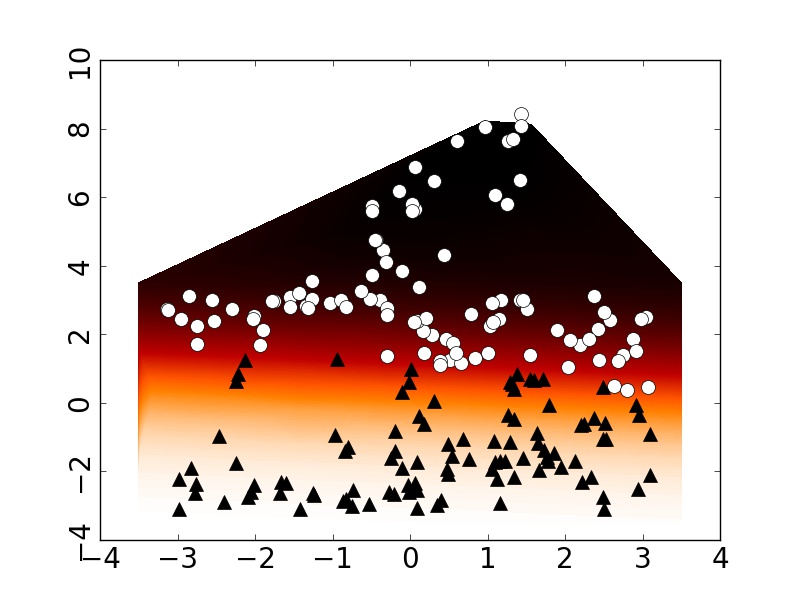}
        \label{fig:noreg300}
    \end{subfigure}
    ~ 
    \begin{subfigure}[b]{0.34\textwidth}
        \includegraphics[width=\textwidth]{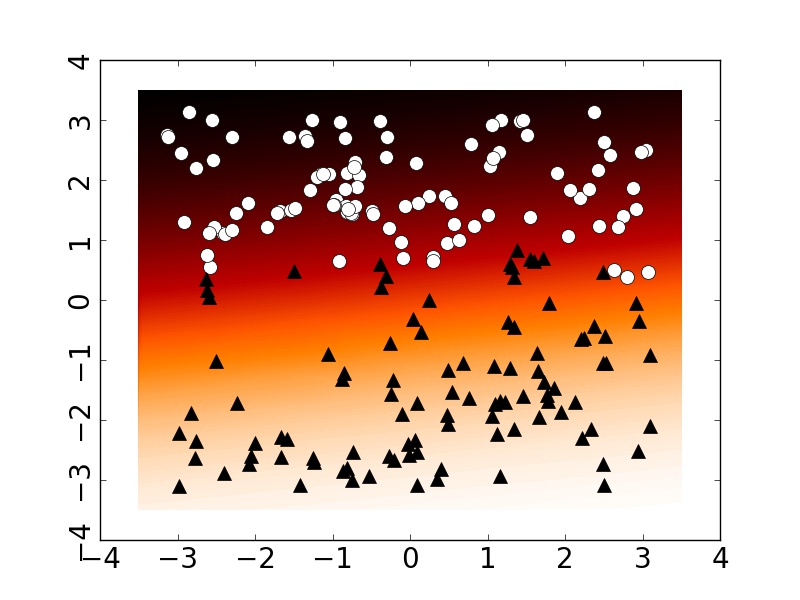}
        \label{fig:reg300}
    \end{subfigure}
    \vspace{-0.25cm}
    \caption{Regularization over \textit{sin} dataset (5000 epochs, $\eta = 3.0$ and $\lambda=0.0005$).}\label{fig:reg}
\end{figure}

In Figure \ref{fig:bound}, it is possible to see that the original boundary layer turns into a hyperplane on the transformed space. Although the algorithm achieved good classification by bending the space and extracting the center of the circle to the outside, it generates a superposition of different points in the original space. Thus, in the region where this happens, misclassification occurs and this should be avoided. One way to diminish the algorithm's power to create such extreme movements is by the use of regularization. Figure \ref{fig:reg} acting as a damper, smoothing the movements happening on the original space, preventing the overlapping of different points in the same region of the transformed space.


The choice of $N =1$ and $h=1$ may facilitate overfitting, since the steps taken by data points are far greater and the constriction on how many movements can be made is at its maximum. It's possible to drawn an analogy of regularization been automatically made when we choose good values of $h$ and $N$, as only small steps are taken and that streamlines are followed.

\section{Conclusion}


This work presents a novel neural network based on vector fields. The base of this network is to move points along a  flow in the space, allowing the posterior separation of these points using a linear classifier. The vector fields are created using kernel functions. This approach brings the vast amount of well established vector fields' theory  enabling  the  geometrical interpretation of how this neural network works. Initialization of parameter and hyperparameter plays an important role on some cases and must be taken into account. Regularization was shown to act as damper, reducing the  capability of the vector field to move points and diminish disruption in the dataset original space. Experiments presented cost function reduction, which indicates learning capability, and the optimized flow used to move the points.

Further work needs to be done exploring real world datasets and evaluating learning performance with validation test sets. Another topic includes the investigation of the architecture performance when hyperparameter $h$ is small and $N$ is large.


\begin{footnotesize}



\bibliographystyle{unsrt}
\bibliography{main.bib}

\begin{thebibliography}{10}

\bibitem{freitas2010importance}
Alex~A. Freitas, Daniela~C. Wieser, and Rolf Apweiler.
\newblock On the importance of comprehensible classification models for protein
  function prediction.
\newblock {\em IEEE/ACM Transactions on Computational Biology and
  Bioinformatics (TCBB)}, 7(1):172--182, 2010.

\bibitem{goyal2016interpreting}
Yash Goyal, Akrit Mohapatra, Devi Parikh, and Dhruv Batra.
\newblock Interpreting visual question answering models.
\newblock In {\em ICML Workshop on Visualization for Deep Learning}, volume~2,
  2016.

\bibitem{barakat2006rule}
Nahla Barakat and Andrew~P. Bradley.
\newblock Rule extraction from support vector machines: Measuring the
  explanation capability using the area under the roc curve.
\newblock In {\em Pattern Recognition, 2006. ICPR 2006. 18th International
  Conference on}, volume~2, pages 812--815. IEEE, 2006.

\bibitem{Freitas2014}
Alex~A. Freitas.
\newblock Comprehensible classification models: A position paper.
\newblock {\em SIGKDD Explor. Newsl.}, 15(1):1--10, March 2014.

\bibitem{zeiler2014visualizing}
Matthew~D. Zeiler and Rob Fergus.
\newblock Visualizing and understanding convolutional networks.
\newblock In {\em European conference on computer vision}, pages 818--833.
  Springer, 2014.

\bibitem{yosinskiunderstanding}
Jason Yosinski, Jeff Clune, Thomas Fuchs, and Hod Lipson.
\newblock Understanding neural networks through deep visualization.
\newblock In {\em Deep Learning Workshop}. 31st International Conference on
  Machine Learning, 2015.

\bibitem{olah2017feature}
Chris Olah, Alexander Mordvintsev, and Ludwig Schubert.
\newblock Feature visualization.
\newblock {\em Distill}, 2017.
\newblock https://distill.pub/2017/feature-visualization.

\bibitem{hornik1991approximation}
Kurt Hornik.
\newblock Approximation capabilities of multilayer feedforward networks.
\newblock {\em Neural networks}, 4(2):251--257, 1991.

\bibitem{colah}
Chris Olah.
\newblock Neural networks, manifolds and topology.
\newblock http://colah.github.io/posts/2014-03-NN-Manifolds-Topology/.

\bibitem{mescheder2017numerics}
Lars Mescheder, Sebastian Nowozin, and Andreas Geiger.
\newblock The numerics of gans.
\newblock {\em arXiv preprint arXiv:1705.10461}, 2017.

\bibitem{Butcher1987}
J.~C. Butcher.
\newblock {\em The Numerical Analysis of Ordinary Differential Equations:
  Runge-Kutta and General Linear Methods}.
\newblock Wiley-Interscience, New York, NY, USA, 1987.

\bibitem{scikit-learn}
F.~Pedregosa, G.~Varoquaux, A.~Gramfort, V.~Michel, B.~Thirion, O.~Grisel,
  M.~Blondel, P.~Prettenhofer, R.~Weiss, V.~Dubourg, J.~Vanderplas, A.~Passos,
  D.~Cournapeau, M.~Brucher, M.~Perrot, and E.~Duchesnay.
\newblock Scikit-learn: Machine learning in {P}ython.
\newblock {\em Journal of Machine Learning Research}, 12:2825--2830, 2011.

\end{thebibliography}

\end{footnotesize}


\end{document}